\documentclass[letterpaper, 10 pt, conference]{ieeeconf}  

\IEEEoverridecommandlockouts                              

\usepackage{graphics} 
\usepackage{epsfig} 
\usepackage{mathptmx} 
\usepackage{times} 
\usepackage{amsmath} 
\usepackage{amssymb}  
\usepackage{booktabs}
\usepackage{multirow}
\usepackage{adjustbox}
\usepackage{array}
\usepackage{makecell}
\usepackage{tablefootnote}
\usepackage{biblatex}
\usepackage{hyperref}

\title{\LARGE \bfseries 
FUSE: Label-Free Image-Event Joint Monocular Depth Estimation via Frequency-Decoupled Alignment and Degradation-Robust Fusion
}

\author{Pihai Sun$^{1,2}$, Junjun Jiang$^{1,*}$, Yuanqi Yao$^{1}$, Youyu Chen$^{1}$, Wenbo Zhao$^{1}$, Kui Jiang$^{1,2}$, Xianming Liu$^{1}$%
\thanks{$^{1}$Faculty of Computing, Harbin Institute of Technology}%
\thanks{$^{2}$Zhengzhou Research Institute, Harbin Institute of Technology}
\thanks{$^{*}$Corresponding author}%
}

\begin{document}

\maketitle

\begin{abstract}
Image-event joint depth estimation methods leverage complementary modalities for robust perception, yet face challenges in generalizability stemming from two factors: 1) limited annotated image-event-depth datasets causing insufficient cross-modal supervision, and 2) inherent frequency mismatches between static images and dynamic event streams with distinct spatiotemporal patterns, leading to ineffective feature fusion. 
To address this dual challenge, we propose Frequency-decoupled Unified Self-supervised Encoder (\textit{FUSE}) with two synergistic components: 

The Parameter-efficient Self-supervised Transfer (PST) leverages image foundation models for cross-modal knowledge transfer, effectively mitigating data scarcity by enabling joint encoding without depth ground truth.

Complementing this, the Frequency-Decoupled Fusion module (FreDFuse) resolves modality-specific frequency mismatches by decoupling features into high- and low-frequency bands and then performing a guided cross-attention fusion, where the modality dominant in each band steers the integration.
This combined approach enables \textit{FUSE} to construct a universal image-event encoder that only requires lightweight decoder adaptation for target datasets. Extensive experiments demonstrate state-of-the-art performance with \(14\%\) and \(24.9\%\) improvements in Abs.Rel on MVSEC and DENSE datasets. 

The framework exhibits remarkable robustness and generalization in challenging scenarios, including extreme lighting and motion blur, significantly advancing its real-world deployment capabilities.
The source code for our method is publicly available at: \url{https://github.com/sunpihai-up/FUSE}.

\end{abstract}


\section{Introduction}

Modern monocular depth estimation (MDE) increasingly utilizes multi-modal sensing to address the limitations of single-sensor systems~\cite{ramnet, xu2024unveiling}, particularly by integrating conventional image sensors with emerging event cameras. Traditional image sensors, which operate using fixed-exposure light integration, effectively preserve structural information but suppress high-frequency temporal variations. This characteristic makes them suitable for static scene perception but limits their performance in dynamic scenarios. In contrast, event cameras asynchronously detect pixel-wise brightness changes with microsecond resolution~\cite{zhu2018ev}, making them highly effective for capturing high-frequency motion but incapable of detecting static structures with subthreshold intensity gradients. The inherent complementarity in the frequency domain between these two modalities highlights the potential of image-event fusion as a promising approach for achieving robust depth estimation in dynamic environments.

\begin{figure}
    \centering
    \includegraphics[width=1\linewidth]{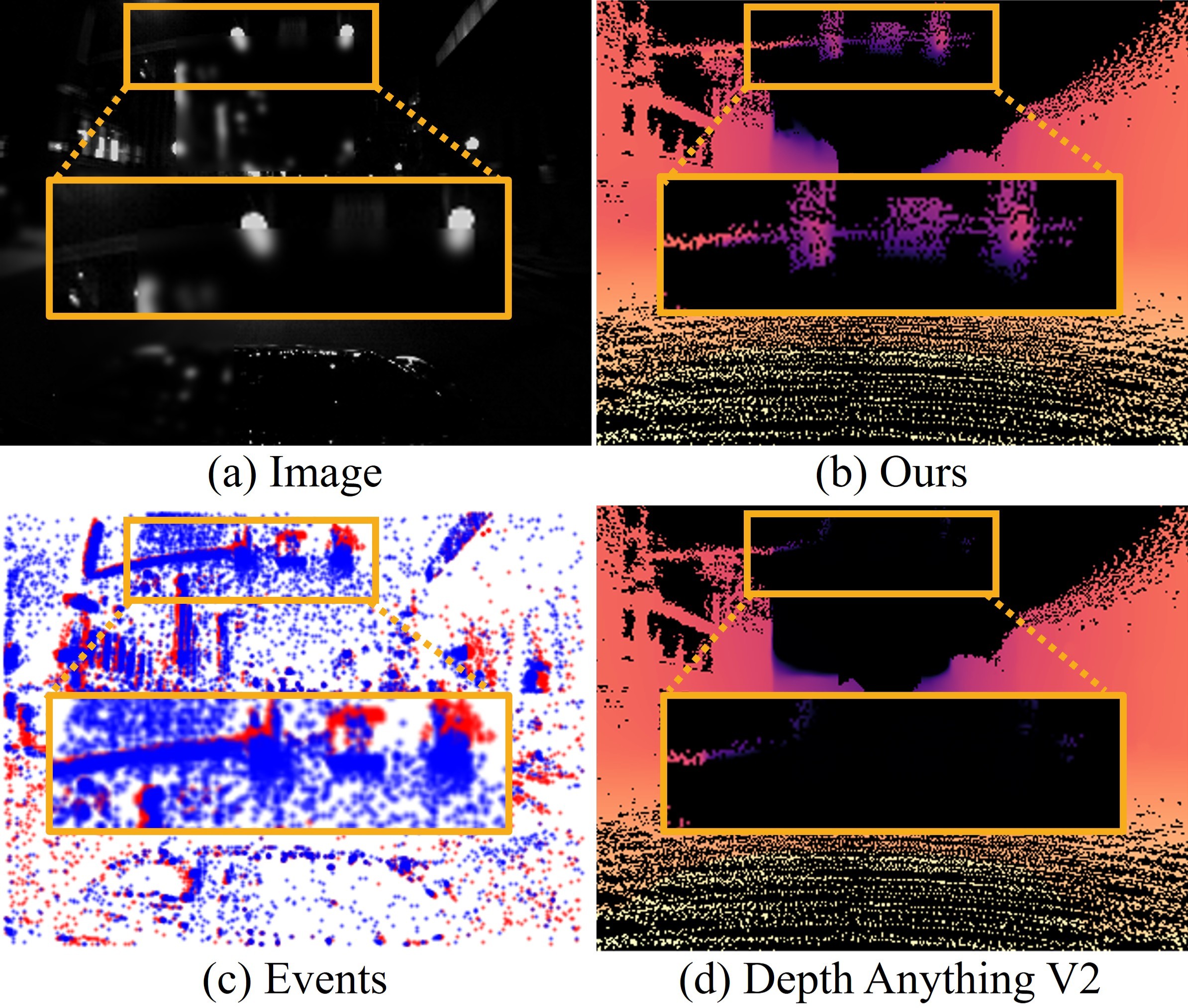}
    \vspace{-20pt}
    \caption{\textbf{Demonstration of \textit{FUSE}'s superior performance in challenging conditions.} (a) and (c) show the input image and event data, with blue/red in (c) indicating brightness decrease/increase. When image data fails due to low light and blur, event data provides complementary dynamic information. Our method (d) leverages multimodal synergy, recovering the traffic light missed by the state-of-the-art image depth model~\cite{depthanything2}, as highlighted in the orange box.}
\label{fig:show}
\vspace{-10pt}
\end{figure}

\begin{figure*}
    \centering
    \includegraphics[width=1\linewidth]{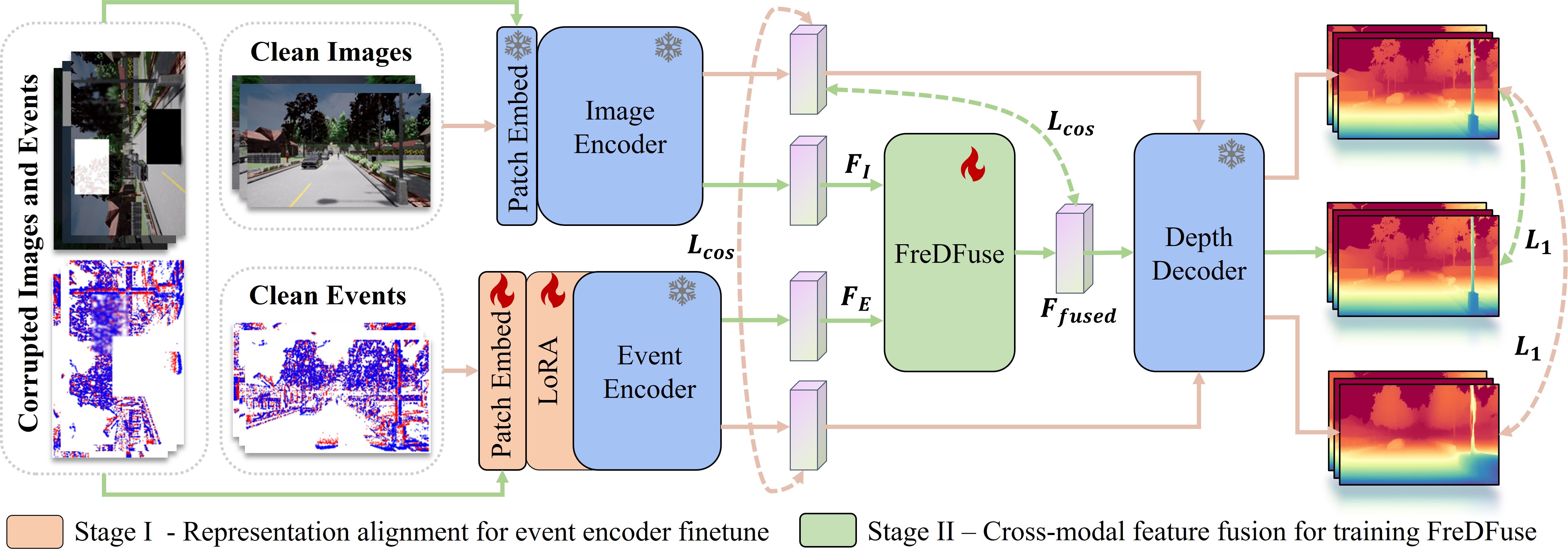}
    \caption{\textbf{Overview of the \textit{FUSE} framework}. \textit{FUSE} integrates an image encoder, event encoder, Frequency-Decoupled Fusion module (FreDFuse), and depth decoder. The image encoder and depth decoder are initialized with a pre-trained MDE model. A two-stage knowledge transfer strategy fine-tunes the event encoder and FreDFuse. In Stage I, the event encoder is initialized with the image encoder’s weights, and only the LoRA matrix and Patch Embed are fine-tuned using clean event data. In Stage 2, randomly degraded image-event pairs are used, and only the FreDFuse is fine-tuned. The MDE model supervises both stages with clean image data in the output and latent spaces.}
    \label{fig:framework}
    \vspace{-10pt}
\end{figure*}

Recent studies~\cite{ramnet, srfnet, pcdepth} have explored leveraging the complementary characteristics of image and event modalities for joint MDE. These methods typically encode both modalities into a unified feature space, followed by customized feature fusion modules for depth prediction. However, their reliance on supervised learning with depth ground truth from image-event pairs has revealed poor generalization capabilities. This limitation arises from the scarcity and limited diversity of existing labeled image-event datasets~\cite{mvsec} compared to large-scale image-depth datasets such as NYUv2~\cite{nyuv2} and KITTI~\cite{kitti}.

Additionally, although the frequency-domain complementarity between images and events presents theoretical advantages, mismatches in their frequency characteristics introduce significant challenges for feature fusion. Conventional fusion strategies risk destructive interference~\cite{chen2024frequency}: high-frequency event features can disrupt the structural continuity of image representations, while low-frequency image components may suppress critical motion cues captured by events. This inherent frequency divergence, combined with limited training data, poses a dual challenge—requiring models to bridge the modality gap with insufficient supervision and resolve conflicting frequency signatures during feature integration.

Inspired by the impressive generalization capabilities of image-based depth foundation models (e.g., Depth Anything~\cite{depthanything1}), which are trained on millions of images, we propose a potential solution: transferring their well-learned image-depth priors to bridge the gap between event-based and depth-based representations. This insight motivates our two-fold innovation: 1) addressing data scarcity through parameter-efficient transfer learning, and 2) resolving intrinsic frequency mismatches via frequency decoupling mechanisms.

We introduce \textit{FUSE}, a frequency-decoupled, unified, self-supervised encoder-based method designed to transfer knowledge from image-based depth models to image-event joint (MDE) without requiring depth ground truth. \textit{FUSE} consists of two main components: 1) Parameter-Efficient Self-Supervised Transfer (PST): A two-phase training strategy with decoupled feature alignment and fusion stages. In Phase I, an event encoder is trained using lightweight adapter tuning to establish a shared latent space between image and event modalities. Phase II focuses on frequency-aware fusion training, incorporating a novel fusion module, FreDFuse, under simulated sensor degradation conditions; 2) Frequency-Decoupled Fusion Module (FreDFuse): Serving as the core of PST's second phase, this module separates features into distinct frequency components before fusion. By utilizing frequency-specific processing pipelines, it enables complementary integration while minimizing inter-frequency interference, thus overcoming the limitations of traditional single-bandwidth fusion approaches.

To validate the effectiveness of \textit{FUSE}, we conduct extensive experiments on the MVSEC and DENSE datasets, demonstrating state-of-the-art (SOTA)  performance in MDE. As shown in Fig. \ref{fig:show}, \textit{FUSE} outperforms the current SOTA image depth model~\cite{depthanything2} under zero-shot settings, particularly in night-time driving scenarios affected by motion blur, highlighting its robustness. In summary, our contributions are as follows:

 \begin{itemize} 
    \item We propose \textit{FUSE}, the first self-supervised framework for generalizable image-event depth estimation that transfers knowledge from image-only depth foundation models without requiring depth-labeled image-event pairs. This addresses the critical challenge of scarce annotated image-event-depth datasets.

    \item Parameter-Efficient Self-Supervised Transfer (PST): A novel two-stage training strategy that bridges the image-event representation gap through lightweight LoRA-based tuning and degradation-robust distillation, significantly reducing trainable parameters compared to full fine-tuning.
    
    \item Frequency-Decoupled Fusion Module (FreDFuse): A frequency-aware fusion module that explicitly disentangles high-frequency edge dynamics (from events) and low-frequency structural priors (from images). This approach resolves modality-specific frequency mismatches while preserving complementary cues.

 \end{itemize}

\section{Related work}

\subsection{Image-based Monocular Depth Estimation}

Deep learning-based methods have become the dominant paradigm in MDE. Eigen et al. \cite{eigen} were the first to introduce a multi-scale fusion network for depth prediction through regression. Subsequent studies have improved depth estimation accuracy by reformulating the regression task as classification~\cite{adabins, binsformer}, introducing novel network architectures \cite{dpt}, and incorporating additional prior knowledge \cite{nddepth, newcrf}. Despite these advancements, the generalization capability of such models remains a major challenge.To address this limitation, recent research~\cite{midas} has focused on training models on large-scale datasets aggregated from diverse sources, leading to substantial improvements. A milestone in this direction is MiDaS \cite{midas}, which mitigates the impact of varying depth scales across datasets by employing an affine-invariant loss. More recently, the DepthAnything series~\cite{depthanything1, depthanything2} has further enhanced the utilization of unlabeled data alongside large-scale labeled datasets, achieving remarkable performance. Concurrently, other research~\cite{yao2024improving} has focused on improving generalization through model-centric approaches, such as employing stabilized adversarial training to boost robustness against domain shifts. However, due to the inherent limitations of traditional image cameras, image-based MDE methods are still susceptible to factors such as lighting conditions and motion blur, which can degrade their performance~\cite{imgbad1}.

\subsection{Event-based Monocular Depth Estimation}

Event-based MDE methods have attracted attention for their effectiveness under extreme lighting and high-speed motion conditions, making them suitable for applications in autonomous driving and robotic navigation \cite{eventsurvey}. E2Depth \cite{e2depth} was the first method to propose dense MDE using event data, employing a recurrent encoder-decoder architecture. Mixed-EF2DNet \cite{mixed} introduced a flow network to capture temporal information effectively, while EReFormer \cite{ereformer} leveraged a Transformer-based architecture for improved performance. Additionally, HMNet \cite{hmnet} incorporated multi-level memory units to handle long-term dependencies efficiently. Nonetheless, these methods continue to face challenges due to the sparsity of event data and the limited availability of training datasets \cite{eventbad}.

\subsection{Image-Event Fusion}
The complementary characteristics of event and image data have been leveraged to enhance various visual tasks, including semantic segmentation \cite{zhang2023cmx, imgbag4}, object detection \cite{zhou2023rgb}, and depth estimation \cite{ramnet, srfnet}. RAMNet \cite{ramnet} employs an RNN-based approach to effectively utilize asynchronous event data, while EVEN \cite{shi2023even} improves low-light performance by fusing modalities prior to inputting them into an MDE network. SRFNet \cite{srfnet} guides modality interaction using spatial reliability masks, and PCDepth \cite{pcdepth} integrates features at the modality level for more effective fusion.
Our \textit{FUSE} is the first to leverage knowledge transfer from an image foundation model for both feature encoding and fusion, significantly enhancing robustness and accuracy in image-event joint depth estimation.

\section{Methodology}
\subsection{Overview}
In this section, we introduce \textit{FUSE}, a low-cost and robust transfer learning framework for generalizable depth prediction.
An overview of the \textit{FUSE} architecture is provided in Fig.~\ref{fig:framework}, our framework consists of four key components: an event encoder \( \mathcal{E}_E \), an image encoder \( \mathcal{E}_I \), a Frequency-Decoupled Fusion module (FreDFuse), and a depth decoder \( \mathcal{D} \).
Given an input image \( \mathbf{I} \) and its event stream \( \mathbf{S} \), the target is to predict the dense depth map \( \mathbf{d}^* \). Our \textit{FUSE} provides a powerful image-event joint encoder (\( \mathcal{E}_I \), \( \mathcal{E}_E \), FreDFuse) with exceptional generalization capabilities, enabling robust performance with minimal adaptation of the depth decoder for target datasets.

In Sec.\ref{sec:event_rep}, we first describe the event processing pipeline. Subsequently, in Sec.\ref{sec:method_lst}, we detail the process of obtaining the image-event joint encoder via PST. Finally, in Sec.~\ref{sec:FreDFuse}, we present the FreDFuse in detail.
\begin{figure}
    \centering
    \includegraphics[width=1\linewidth]{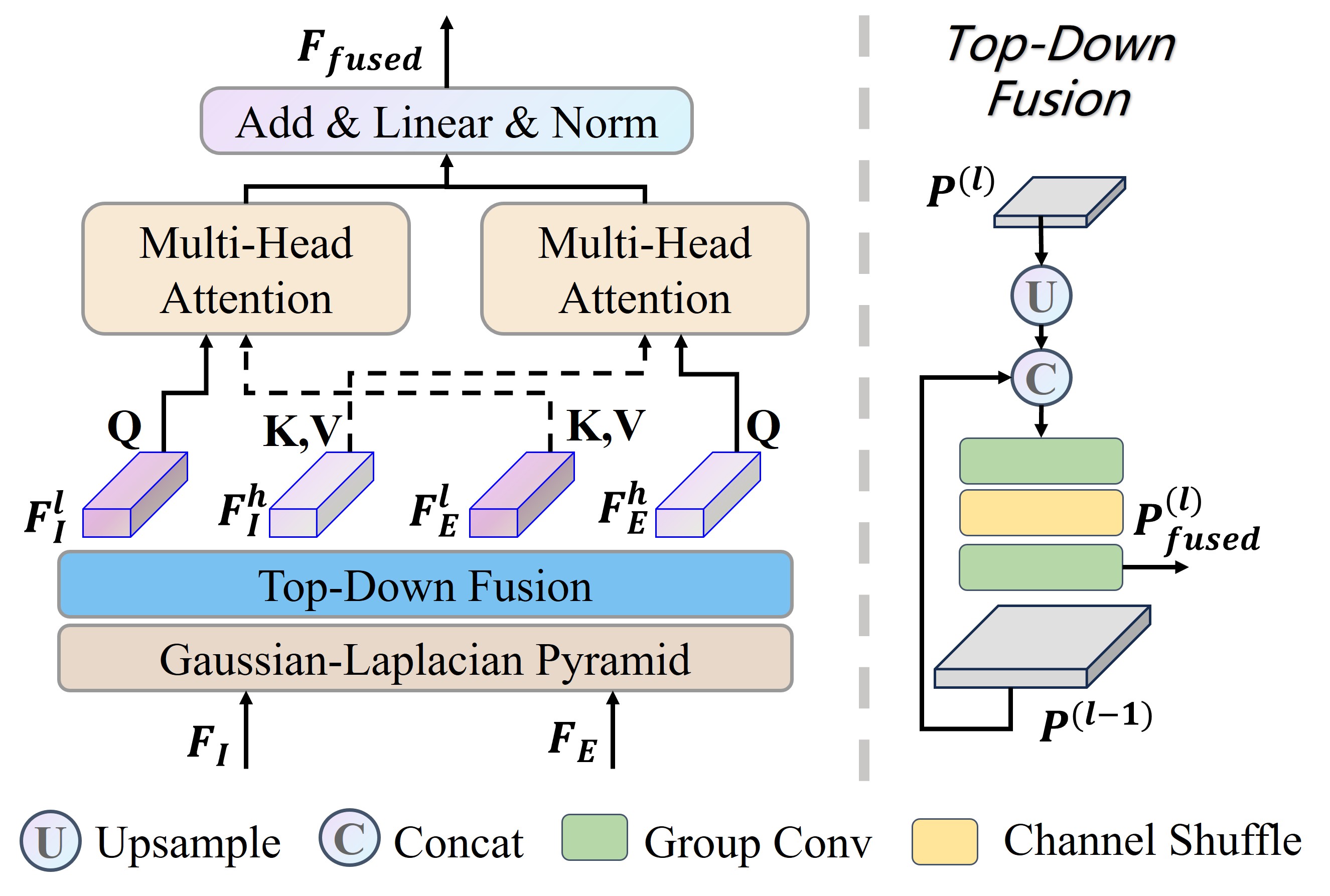}
    \caption{\textbf{Overview of our FreDFuse.} FreDFuse decouples image features \(\mathbf{F}_I\) and event features \(\mathbf{F}_E\) into high- and low-frequency components using a Gaussian-Laplacian pyramid. Multi-scale features are fused top-down with \(1 \times 1\) group convolutions and channel shuffle. Fusion in the high-frequency branch is event-driven, while the low-frequency branch is image-driven. The high- and low-frequency components are combined through addition and processed by LayerNorm.}
    \label{fig:FreDFuse}
    \vspace{-10pt}
\end{figure}

\begin{table*}[]
\renewcommand\arraystretch{1.2}
\centering
\caption{Quantitative results on the MVSEC datast~\cite{mvsec}. \(\downarrow\) indicates lower is better and denotes higher is better. The \textbf{best results} are highlighted in bold, while the \underline{second-best outcomes} are underlined.}

\label{tab:expmvsec}
\begin{adjustbox}{width=\textwidth}
\begin{tabular}{cc|cccccc|cccccc}
\Xhline{1.1pt}
\multirow{2}{*}{Methods} & \multirow{2}{*}{Input} & \multicolumn{6}{c|}{outdoor day1}                                                                   & \multicolumn{6}{c}{outdoor night1}                                                                  \\
                  &                        & \(\delta_1 \uparrow\) & \(\delta_2 \uparrow\) & \(\delta_3 \uparrow\) & Abs.Rel\(\downarrow\) & RMSE\(\downarrow\) & RMSELog\(\downarrow\) & \(\delta_1 \uparrow\) & \(\delta_2 \uparrow\) & \(\delta_3 \uparrow\) & Abs.Rel\(\downarrow\) & RMSE\(\downarrow\) & RMSELog\(\downarrow\) \\ \hline
E2Depth\cite{e2depth}                  & E                      & 0.567          & 0.772          & 0.876          & 0.346          & 8.564          & 0.421          & 0.408          & 0.615          & 0.754          & 0.591          & 11.210         & 0.646          \\
EReFormer\cite{ereformer}                & E                      & 0.664          & 0.831          & 0.923          & 0.271          & -              & 0.333          & 0.547          & 0.753          & 0.881          & 0.317          & -              & 0.415          \\
HMNet\cite{hmnet}                    & E                      & 0.690          & 0.849          & 0.930          & 0.254          & 6.890          & 0.319          & 0.513          & 0.714          & 0.837          & 0.323          & 9.008          & 0.482          \\ \hline
RAMNet\cite{ramnet}                   & I+E                    & 0.541          & 0.778          & 0.877          & 0.303          & 8.526          & 0.424          & 0.296          & 0.502          & 0.635          & 0.583          & 13.340         & 0.830          \\
SRFNet\cite{srfnet}                   & I+E                    & 0.637          & 0.810          & 0.900          & 0.268          & 8.453          & 0.375          & 0.433          & 0.662          & 0.800          & 0.371          & 11.469         & 0.521          \\
HMNet\cite{hmnet}                    & I+E                    & {\underline{0.717}}    & {\underline{0.868}}    & 0.940          & 0.230          & 6.922          & 0.310          & 0.497          & 0.661          & 0.784          & 0.349          & 10.818         & 0.543          \\
PCDepth\cite{pcdepth}                  & I+E                    & 0.712          & 0.867          & {\underline{0.941}}    & {\underline{0.228}}    & {\underline{6.526}}    & {\underline{0.301}}    & \textbf{0.632} & {\underline{0.822}}    & {\underline{0.922}}    & {\underline{0.271}}    & {\underline{6.715}}    & {\underline{0.354}}    \\ \hline
\textbf{\textit{FUSE}}                     & I+E                    & \textbf{0.745} & \textbf{0.892} & \textbf{0.957} & \textbf{0.196} & \textbf{6.004} & \textbf{0.270} & {\underline{0.629}}    & \textbf{0.824} & \textbf{0.923} & \textbf{0.261} & \textbf{6.587} & \textbf{0.351} \\ \Xhline{1.1pt}
\end{tabular}
\end{adjustbox}
\end{table*}

\subsection{Event Representation}\label{sec:event_rep}
The event stream is first transformed into a voxel grid representation~\cite{ramnet} to make it compatible with existing network architectures. 
Each event \( e_n \) is represented by the tuple \( (x_n, y_n, t_n, p_n) \), which means that at time \( t_n \), the log intensity change at the pixel position \( (x_n, y_n) \) exceeds a threshold. The polarity \( p_n \ (\pm 1) \) indicates whether the intensity increases or decreases. As such, the asynchronous event stream \( \mathbf{S} = \{ e_n \}_{n=0}^{N-1} \) must be transformed into a frame-like representation to be compatible with existing network architectures. Following prior works \cite{ramnet}, \cite{srfnet}, \cite{pcdepth}, \cite{e2depth} we convert the temporal stream of events into a voxel grid \cite{gehrig2019end}.

Given an event stream \( \mathit{S} = \{ e_n \}_{n=0}^{N-1} \) within the time interval \( \Delta T = t_{N-1} - t_0 \), we transform it into a voxel grid with spatial dimensions \( H \times W \) and \( B \) time bins:
\begin{equation}
\mathit{V}(x, y, t) = \sum p_i \delta\left(x - x_i, y - y_i \right) \max \left\{ 0, 1 - \left| t - t_i^* \right| \right\},
\end{equation}
where \( t_i^* = \frac{B-1}{\Delta T} (t_i - t_0) \). The voxel grid \( \mathit{V} \in \mathbb{R}^{H \times W \times B} \). We choose \( B = 3 \) time bins.

The similar structural form facilitates knowledge transfer from image-based models, enabling the leveraging of pre-trained networks for enhanced performance. At the same time, the image data can be used without additional processing. 

\subsection{Parameter-efficient Self-supervised Transfer}
\label{sec:method_lst}

Our Parameter-efficient Self-supervised Transfer (PST) strategy addresses the critical challenge of event-based depth estimation data scarcity by establishing a unified foundation model for image-event joint MDE through efficient cross-modal knowledge transfer. This self-supervised approach eliminates dependency on expensive depth ground truth while achieving two key objectives: 1) transferring image domain priors to the image-event joint domain using only paired image-event data, and 2) preserving the foundation model's generalization capability through parameter-efficient adaptation. As shown in Fig.~\ref{fig:framework}, PST operates through two cascaded stages:

\noindent \textbf{Stage I: Parameter-Efficient Feature Alignment}:
\label{sec:stage1}
In this stage, we align event data representations with the image domain's latent space. We initialize \(\mathcal{E}_E\) and \(\mathcal{D}_E\) using weights 
from Depth Anything V2~\cite{depthanything2} to inherit its geometric understanding ability. 
To enable efficient adaptation while preventing catastrophic forgetting \cite{forgetcatastrophic}, we implement LoRA\cite{lora}, where only the LoRA matrices and PatchEmbed Layer parameters (1.8\% of \(\mathcal{E}_E\)) are updated. This allows adaptation to event data while retaining the foundation model's generalization capability. 
The alignment is driven by a compound objective function (detailed in Sec.~\ref{sec:loss}) that enforces consistency in both output space (depth prediction) and latent representations.

\noindent \textbf{Stage II: Robust Feature Fusion}: 
This stage focuses on training only the FreDFuse module (Sec.~\ref{sec:FreDFuse}) for robust multi-modal fusion, while keeping other components frozen: the image encoder ($\mathcal{E}_I$) and decoder ($\mathcal{D}$) from a foundation model (e.g., Depth Anything V2), and the event encoder ($\mathcal{E}_E$) from Stage I. 
To improve robustness and generalization, we employ a systematic data augmentation strategy by training on randomly degraded image-event pairs. 
This includes global brightness adjustments (factor $\in [0.5, 1.5]$) and localized degradations applied to 50\% of the pairs in random $20\% \times 20\%$ regions (detailed in Table~\ref{tab:degradation_params}). 
This strategy forces the fusion module to learn resilient cross-modal correlations, using the objective from Eq.~\ref{lstloss} to enforce consistency.

\begin{table}[htbp]
    \centering
    \caption{Parameterization of the random degradation strategy.}
    \label{tab:degradation_params}
    \begin{adjustbox}{width=\linewidth}
    \resizebox{\columnwidth}{!}{%
    \begin{tabular}{@{}c|cc@{}}
        \Xhline{1.1pt}
        Degradation Type & Parameters & Applicable Modality \\ \hline
        Gaussian Blur & Kernel size: $25 \times 25$, $\sigma=2$ & Image \& Event \\
        Overexposure  & $\alpha=2.5$, $\beta=200$             & Image only    \\
        Occlusion     & Fill region with black pixels         & Image \& Event \\
        \Xhline{1.1pt}
    \end{tabular}
    }
    \end{adjustbox}
\end{table}

\subsection{Frequency-Decoupled Fusion Module}
\label{sec:FreDFuse}
To leverage the complementary characteristics of different visual sensors, we propose a frequency-decoupled fusion method. 
Image sensors excel at capturing low-frequency static scenes, while event cameras are adept at detecting high-frequency dynamic changes. Our Frequency-Decoupled Feature Integration module (FreDFuse) leverages these complementary strengths by processing and fusing features in separate frequency domains, enabling more effective multi-modal depth estimation. Fig.~\ref{fig:FreDFuse} shows the overview of FreDFuse.

\noindent \textbf{Frequency Decoupling via Gaussian-Laplacian Pyramids}: 
Given token sequences from image features \( \mathbf{F}_I \in \mathbb{R}^{B \times N \times C} \) and event features \( \mathbf{F}_E \in \mathbb{R}^{B \times N \times C} \), we first decompose them into low-frequency and high-frequency components using Gaussian-Laplacian pyramids. For each modality \( m \in \{I, E\} \), the decomposition is formulated as: 

\begin{equation}
\mathbf{G}_m, \mathbf{L}_m = \mathcal{P}_m(\mathbf{F}_m),
\end{equation} 
where \( \mathcal{P}_m(\cdot) \) denotes the Gaussian-Laplacian decoupling operator. Specifically, the input tokens are reshaped into 2D feature maps and processed through multi-scale pyramids. The Gaussian pyramid \( \mathbf{G}_m \) captures low-frequency components via iterative blurring and downsampling:  

\begin{equation}
\mathbf{G}_m^{(l)} = \text{Down}(\text{GaussianBlur}(\mathbf{G}_m^{(l-1)})),
\end{equation} 
where \( l \in \{1, \dots, L\}, L=3 \) denotes the pyramid level. The Laplacian pyramid \( \mathbf{L}_m \), encoding high-frequency details, is derived by subtracting upsampled Gaussian levels:  

\begin{equation}
\mathbf{L}_m^{(l)} = \mathbf{G}_m^{(l)} - \text{Up}(\mathbf{G}_m^{(l+1)}).
\end{equation} 

\noindent \textbf{Top-Down Fusion of Multi-Level Pyramid Features}: 
For each pyramid (\(\mathbf{G}_I, \mathbf{L}_I, \mathbf{G}_E, \mathbf{L}_E\)), multi-scale features are fused by upsampling and concatenating layers, followed by grouped convolutions:

\begin{equation}
\mathbf{P}_{\text{cat}}^{(l)} = \text{Concat}(\mathbf{P}^{(l-1)}, \text{Up}(\mathbf{P}^{(l)})),
\end{equation}
\begin{equation}
\mathbf{P}_{\text{comp}}^{(l)} = \text{GroupConv}_{1 \times 1}(\mathbf{P}_{\text{cat}}^{(l)}),
\end{equation}
\begin{equation}
\mathbf{P}_{\text{fused}}^{(l)} = \text{GroupConv}_{1 \times 1}(\text{ChannelShuffle}(\mathbf{F}_{\text{comp}}^{(l)})).
\end{equation}

\noindent \textbf{Cross-Branch Frequency-Guided Fusion}: 
The decoupled features are fused in two branches, each guided by the dominant modality in the target frequency band. For the high-frequency branch, \( \mathbf{F}_E^{\text{high}} \) is the Query, while \( \mathbf{F}_I^{\text{high}} \) is the Key and Value. For the low-frequency branch, \( \mathbf{F}_I^{\text{low}} \) is the Query, while \( \mathbf{F}_E^{\text{low}} \) is the Key and Value.
Both branches use multi-head cross-attention:
\begin{equation}
\text{Attn}(\mathbf{Q}, \mathbf{K}, \mathbf{V}) = \text{Softmax}\left(\frac{\mathbf{Q} \mathbf{K}^T}{\sqrt{d}}\right) \mathbf{V},
\end{equation}
where \( d \) is the dimension of each attention head. The fused features are recombined and projected to the original space:
\begin{equation}
\mathbf{F}_{\text{fused}} = \text{LayerNorm}\left(\text{Linear} (\mathbf{F}_{\text{low}} + \mathbf{F}_{\text{high}})\right).
\end{equation}

\subsection{Optimization Objectives}

\noindent \textbf{Optimization Objective of PST}:
\label{sec:loss}
The PST framework adopts a compound loss to achieve two critical objectives: 1) prediction-level distillation via output-space alignment, and 2) latent-space regularization to preserve transferable representations.
Given clean image \(\mathbf{I}\), the image foundation model generates pseudo depth \(\mathbf{d}\) and latent features \(\mathbf{F}\), which provide supervision for Stage I and Stage II. The depth map and latent features of the output of Stage I and Stage II are defined as \(\mathbf{d^*}\) and \(\mathbf{F}^*\). We formulate a compound loss function \({L}_{\text{align}}\):
\begin{equation} \label{lstloss}
    {L}_{\text{align}} = {L}_1 + {L}_{\cos},
\end{equation}
where \({L}_1 = ||{d} - {d}^*||_1\) aligns output predictions, and \({L}_{\cos} = \left[1 - \cos(\mathbf{F}, \mathbf{F}^*)\right] \cdot \mathbb{I}({\alpha \leq \cos(\mathbf{F}, \mathbf{F}^*) \leq \beta})\) aligns latent representations. Following the setup in related work~\cite{depthanything1}, we set \(\alpha = 0.2\) and \(\beta = 0.85\) to prevent over-regularization, where \(\mathbb{I}(\cdot)\) ensures loss activation only within the specified similarity range.

\noindent \textbf{Optimization Objective of Target Datasets}:
When transferring to target datasets with metric depth labels, we replace pseudo-supervision with the Scale-invariant Logarithmic (SiLog) loss~\cite{eigen}. Given a predicted depth map \(\mathbf{d^{*}}\) and depth ground truth \(\mathbf{d}\), the loss is computed over the pixels as follows: 
\begin{equation}
{L}_{\text{SiLog}} = \sqrt{ \frac{1}{N} \sum_{i} e_i^2 - \lambda \left( \frac{1}{N} \sum_{i} e_i \right)^2 },
\end{equation}
where \(e_i = \ln d_i - \ln d^{*}_i\), \(N\) is the total number of valid pixels. 

\begin{figure*}
    \centering
    \includegraphics[width=1\linewidth]{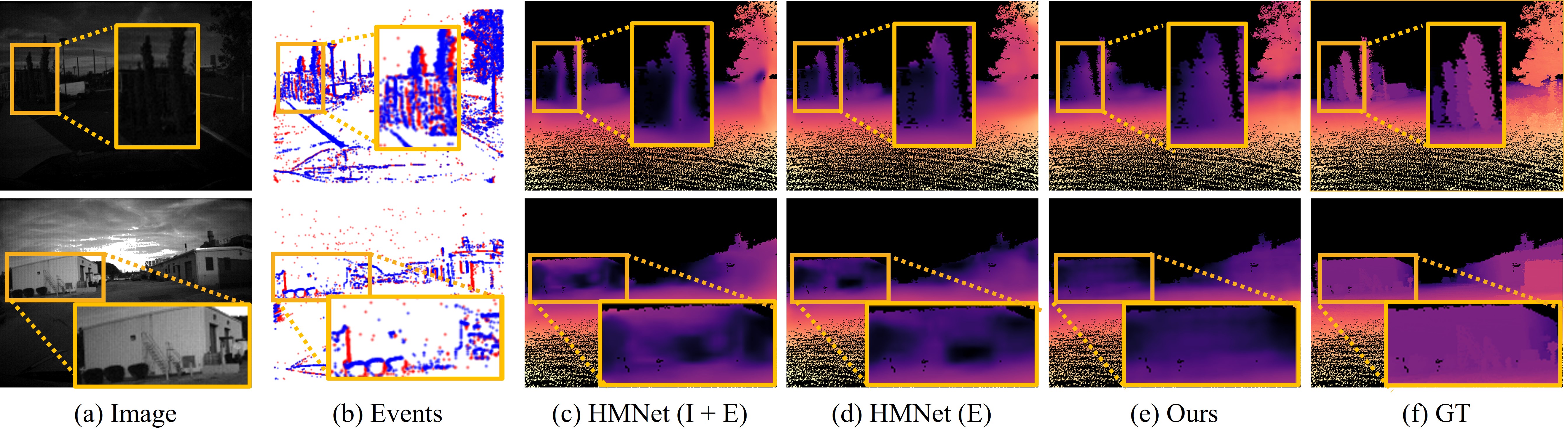}
    \vspace{-20pt}
    \caption{\textbf{Qualitative analysis of the MVSEC dataset, outdoor\_day1 scene.} (a) and (b) show the input image and event data; (c) depicts the image-event joint estimation by HMNet\cite{hmnet}; (d) shows the event-only estimation by HMNet; (e) presents our proposed \textit{FUSE}; (f) shows the depth ground truth.}
    \label{fig:visday1}
\end{figure*}

\begin{figure*}
    \centering
    \includegraphics[width=1\linewidth]{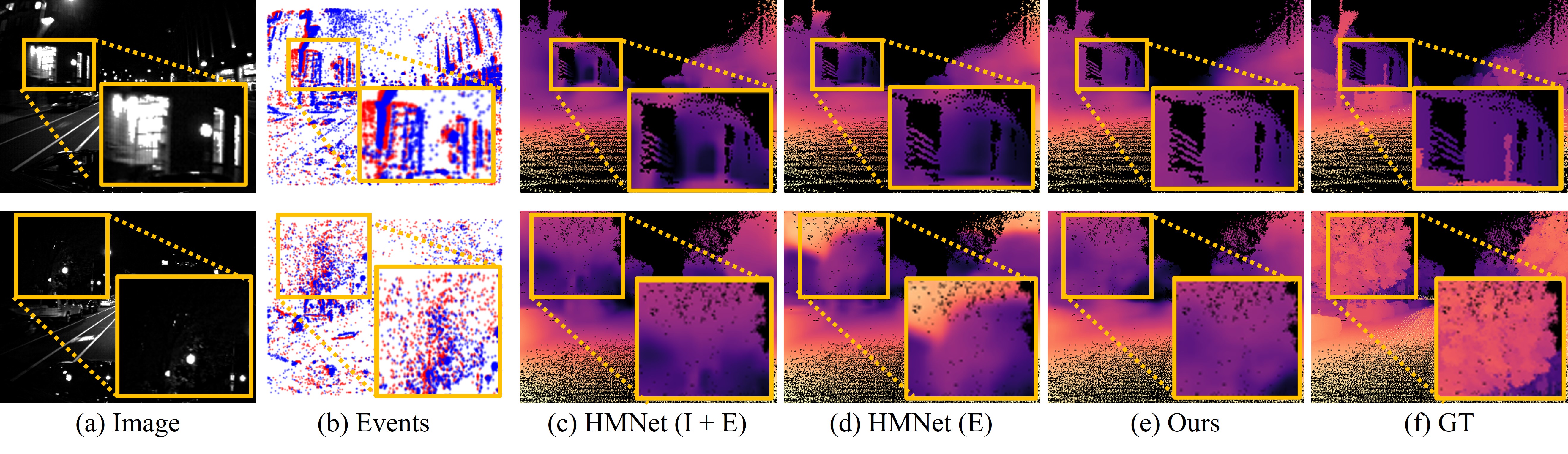}
    \vspace{-20pt}
    \caption{\textbf{Qualitative analysis of the MVSEC dataset, outdoor\_night1 scene.} (a) and (b) show the input image and event data; (c) depicts the image-event joint estimation by HMNet\cite{hmnet}; (d) shows the event-only estimation by HMNet; (e) presents our proposed \textit{FUSE}; (f) shows the depth ground truth.}
    \label{fig:visnight1}
\end{figure*}

\begin{figure*}
    \centering
    \includegraphics[width=1\linewidth]{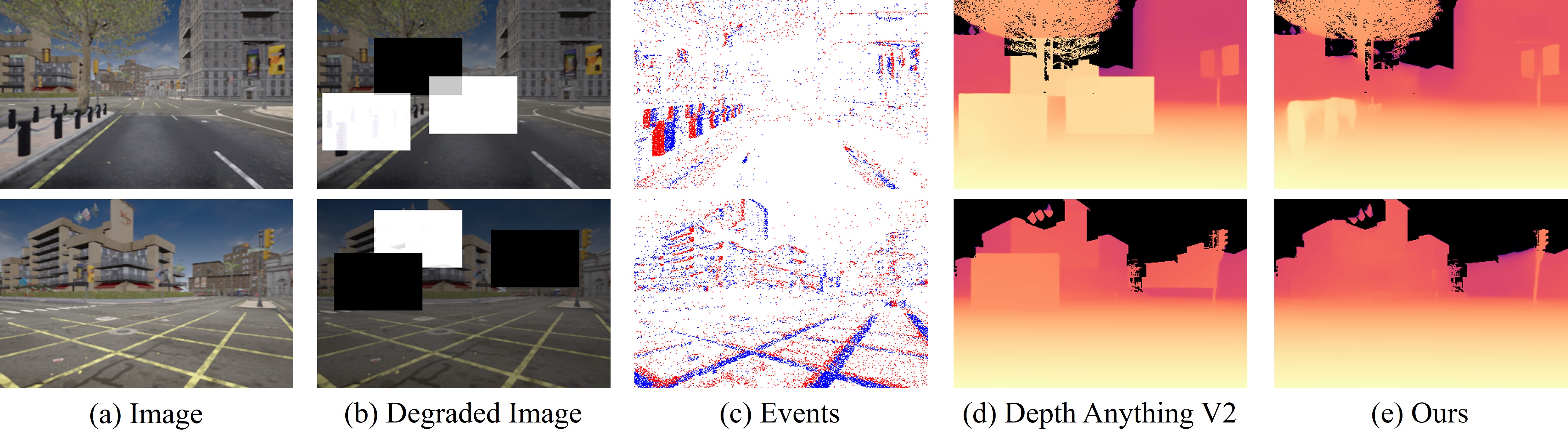}
    \vspace{-20pt}
    \caption{\textbf{Qualitative comparison of zero-shot predictions when the image modality is severely degraded.}(a) denotes the original image, (b) represents the image degraded by low light, overexposure, and occlusion masks, (c) denotes the event data, and (d) and (e) show the predicted results from DepthAnything V2~\cite{depthanything2} and our \textit{FUSE}, respectively.}
    \label{fig:zero}
    \vspace{-10pt}
\end{figure*}


\section{Experiment}
To validate the effectiveness of our proposed \textit{FUSE} framework in addressing the scarcity of event-depth data and modality disparity, we conduct comprehensive experiments under strictly controlled conditions. 

\subsection{Experimental Settings}
We utilize EventScape \cite{ramnet} for image-event pairs in PST (without depth ground truth). We evaluate on MVSEC \cite{mvsec} and DENSE \cite{e2depth} datasets, where MVSEC is a real-world scenario while EventScape and DENSE are synthetic scenarios.

\noindent\textbf{Metrics}: Following \cite{pcdepth}, we evaluate performance using the metrics: Absolute Relative Error (Abs.Rel), Root Mean Square Error (RMSE), Logarithmic Squared Error (RMSELog), average absolute depth errors at different cut-off depth distances (i.e., 10m, 20m, and 30m), and accuracy (\(\delta < 1.25^{n}, n=1,2,3\)).

\noindent\textbf{Implementation Details}: 
The model is trained using the Adam optimizer with a learning rate of 5e-5, implemented in PyTorch on two RTX 3090ti GPUs with a batch size of 48. PST on EventScape~\cite{ramnet} is trained for 10 epochs, and fine-tuning on MVSEC and DENSE for 20 epochs. Manual alignment is needed for the MVSEC dataset due to the asynchronous nature of images, events, and depth labels. As in \cite{pcdepth}, event data from the first 50 ms prior to each depth ground truth is paired with the most recent image to form the data pairs.

\subsection{Evaluation on MVSEC and DENSE Datasets}
\label{sec:exp}

We evaluate our method on the real-world MVSEC \cite{mvsec} and synthetic DENSE \cite{e2depth} datasets. For MVSEC, we train on outdoor\_day2, outdoor\_night2, and outdoor\_night3, and test on outdoor\_day1 and outdoor\_night1 (depth range: 1.97-80m) \cite{pcdepth}. For DENSE, we use its official splits (depth range: 3.34-1000m) \cite{ramnet, srfnet}. During training, we freeze the joint encoder and only train the depth decoder.

\noindent\textbf{Analysis on MVSEC Dataset}: 
Tab. \ref{tab:expmvsec} presents a quantitative comparison between our method and SOTA methods for event-based estimation and image-event joint estimation on the MVSEC dataset. 
In the outdoor\_day1 scene, our method outperforms the best image-event joint estimation approach \cite{pcdepth} by \(14\%\) and \(10.2\%\) in Abs.Rel and RMSELog, respectively. The progress achieved by freezing the encoder and dealing with the significant disparity between source and target data clearly demonstrates the effectiveness of our proposed \textit{FUSE} in addressing the scarcity of event depth data and providing robust and accurate depth estimation. 
Fig. \ref{fig:visday1} and Fig. \ref{fig:visnight1} present qualitative comparisons with HMNet\cite{hmnet} for the outdoor\_day1 and outdoor\_night1 scenes, respectively. \textit{FUSE} provides finer and more stable depth predictions. Compared to HMNet, our method demonstrates a superior ability to preserve structural details, delivering more consistent and accurate predictions on both buildings and vegetation.

\noindent\textbf{Analysis on DENSE Dataset}: Tab. \ref{tab:expdense} presents the quantitative results on the synthetic DENSE dataset \cite{e2depth}. Our method achieves improvements of \(24.9\%\) in Abs.Rel and \(33.4\%\) in RMSELog over previous approaches. For the average absolute depth error at the truncated distance, our method ranks either the best or the second-best among the evaluated methods. Fig.~\ref{fig:zero} shows a qualitative comparison under extreme damage to the image modality. Our \textit{FUSE} is still able to provide stable depth estimation even when some regions of the image are completely destroyed.

\begin{table}[]
    \renewcommand\arraystretch{1.2}
    \caption{Comparisons on synthetic DENSE dataset~\cite{e2depth}}
    \label{tab:expdense}
    \begin{adjustbox}{width=\linewidth}
    \resizebox{\columnwidth}{!}{%
    \begin{tabular}{cc|cc|ccc}
    \Xhline{1.1pt}
    \multirow{2}{*}{Methods} & \multirow{2}{*}{Input} & \multirow{2}{*}{Abs.Rel\(\downarrow\)} & \multirow{2}{*}{RMSELog\(\downarrow\)} & \multicolumn{3}{c}{Avg.Error\(\downarrow\)}      \\
                             &                        &                                        &                                        & 10m            & 20m            & 30m            \\ \hline
    RAMNet\cite{ramnet}                   & I+E                    & 1.189                                  & 0.832                                  & 2.619          & 11.264         & 19.113         \\
    SRFNet\cite{srfnet}                   & I+E                    & \underline{0.513}                      & \underline{0.687}                      & \underline{1.503}          & \textbf{3.566}          & \textbf{6.116}          \\ \hline
    \textbf{\textit{FUSE}}             & I+E                    & \textbf{0.385}                          & \textbf{0.457}                         & \textbf{1.286} & \underline{3.998} & \underline{6.639} \\ \Xhline{1.1pt}
    \end{tabular}%
    }
    \end{adjustbox}
    \end{table}

\subsection{Ablation Studies}

\begin{table}[]
\caption{Ablation studies for Image-Event fusion models}
\label{tab:ablationset}
\begin{adjustbox}{width=\linewidth}
\resizebox{\columnwidth}{!}{%
\begin{tabular}{c|ccc}
\Xhline{1.1pt}
           & Knowledge Transformer & Fusion Module & Paras (M) \\ \hline
Baseline-1 & None                  & Cross-Attention       & 49.2                \\
Baseline-2 & None                  & FreDFuse                 & 51.6                \\
Baseline-3 & One-stage             & FreDFuse                 & 6+2.7               \\ \hline
\textbf{\textit{FUSE}}      & PST                   & FreDFuse                 & 1.4+4.7+2.7        \\ \Xhline{1.1pt}
\end{tabular}%
}
\end{adjustbox}
\end{table}

We conduct ablation studies on the MVSEC dataset using the ViT-Small backbone (Tab.~\ref{tab:ablationset}). Baseline-1 and Baseline-2 are trained from scratch with depth supervision; the former uses cross-attention for fusion, while the latter adopts our proposed FreDFuse (Sec.~\ref{sec:FreDFuse}). Baseline-3 adds one-stage knowledge transfer by initializing the event encoder with image-based foundation model weights and training it jointly with FreDFuse. Tab.~\ref{tab:ablation_day} and Tab.~\ref{tab:ablation_night} report results on outdoor\_day1 and outdoor\_night1. Both PST and FreDFuse contribute significantly to performance gains.

\noindent \textbf{Effectiveness of FreDFuse:} 
In the outdoor\_day1 scene, Baseline-2 demonstrates an \(8\%\) improvement over Baseline-1 in the Abs. Rel metric. Similar improvements are observed across other scenes and metrics. This performance gain can be attributed to FreDFuse's ability to decouple features into high-frequency and low-frequency branches. By allowing the modality that excels in each branch to dominate the feature fusion, FreDFuse effectively mitigates destructive interference caused by frequency mismatches and enhances inter-modal complementation.

\noindent \textbf{Effectiveness of PST:} 
Compared to Baseline-2, \textit{FUSE} integrates our proposed PST, which reduces the training parameters by \(82.2\%\) while achieving an average performance improvement of \(19.7\%\) across all metrics and scenarios. This finding suggests that PST can effectively leverage the knowledge from image-based foundation models to mitigate the data scarcity issue in image-event joint depth estimation. In comparison to Baseline-3, \textit{FUSE}, which employs the complete two-stage knowledge transfer process, also shows performance improvements on most metrics in various scenarios. This highlights the significance of the two-stage training with degraded image-event pairs for enhancing the model's generalization and robustness.

\begin{table}[!t]
    \renewcommand\arraystretch{1.2}
    \caption{Quantitative results of the ablation experiments on the MVSEC \cite{mvsec} outdoor\_day1}
    \label{tab:ablation_day}
    \resizebox{\columnwidth}{!}{%
    \begin{tabular}{c|cccc|ccc}
    \Xhline{1.1pt}
    \multirow{2}{*}{Methods} & \multirow{2}{*}{\(\delta_1\) \(\uparrow\)} & \multirow{2}{*}{Abs.Rel\(\downarrow\)} & \multirow{2}{*}{RMSE\(\downarrow\)} & \multirow{2}{*}{RMSELog\(\downarrow\)} & \multicolumn{3}{c}{Avg,Error\(\downarrow\)}                  \\
                             &                                 &                                        &                                     &                                        & 10m            & 20m                  & 30m                  \\ \hline
    Baseline-1               & 0.612                           & 0.366                                  & 8.451                               & 0.453                                  & 1.711          & 2.754                & 3.303                \\
    Baseline-2               & 0.633                           & 0.336                                  & 8.403                               & 0.433                                  & 1.488          & 2.519                & 3.137                \\
    Baseline-3               & {\underline{0.704}}                     & {\underline{0.233}}                            & {\underline{6.690}}                         & {\underline{0.313}}                            & {\underline{1.076}}    & \textbf{1.802}       & \textbf{2.210}       \\ \hline
    \textbf{\textit{FUSE}}                    & \textbf{0.719}                  & \textbf{0.229}                         & \textbf{6.169}                      & \textbf{0.294}                         & \textbf{1.035} & {\underline{{1.848}}} & {\underline{{2.267}}} \\ \Xhline{1.1pt}
    \end{tabular}%
    }
    \end{table}
    
    \begin{table}[!t]
    \renewcommand\arraystretch{1.2}
    \caption{Quantitative results of the ablation experiments on the MVSEC \cite{mvsec} outdoor\_night1}
    \label{tab:ablation_night}
    \resizebox{\columnwidth}{!}{%
    \begin{tabular}{c|cccc|ccc}
    \Xhline{1.1pt}
    \multirow{2}{*}{Methods} & \multirow{2}{*}{\(\delta_1\)\(\uparrow\)} & \multirow{2}{*}{Abs.Rel\(\downarrow\)} & \multirow{2}{*}{RMSE\(\downarrow\)} & \multirow{2}{*}{RMSELog\(\downarrow\)} & \multicolumn{3}{c}{Avg,Error\(\downarrow\)}      \\
                             &                                 &                                        &                                     &                                        & 10m            & 20m            & 30m            \\ \hline
    Baseline-1               & 0.525                           & 0.356                                  & 7.698                               & 0.413                                  & 1.899          & 2.634          & 3.077          \\ 
    Baseline-2               & 0.531                           & 0.351                                  & 7.583                               & 0.418                                  & 1.817          & 2.545          & 3.018          \\
    Baseline-3               & {\underline{0.611}}                     & {\underline{0.268}}                            & {\underline{6.873}}                         & {\underline{0.354}}                            & {\underline{1.326}}    & {\underline{2.025}}    & {\underline{2.543}}    \\ \hline
    \textbf{\textit{FUSE}}                    & \textbf{0.613}                  & \textbf{0.267}                         & \textbf{6.785}                      & \textbf{0.352}                         & \textbf{1.305} & \textbf{1.998} & \textbf{2.513} \\ \Xhline{1.1pt}
    \end{tabular}%
    }
    \end{table}

\subsection{Computational Overhead Analysis}

We analyze the computational overhead of our \textit{FUSE} framework in terms of parameters, GFLOPs, and inference latency, using a ViT-L backbone and $266 \times 266$ input resolution on a single RTX 3090 Ti GPU. We compare three models: (1) an image-only baseline (Depth Anything V2), (2) a dual-stream baseline with standard cross-attention (Baseline-1), and (3) our \textit{FUSE} with FreDFuse.

As shown in Table~\ref{tab:overhead}, \textit{FUSE} roughly doubles the parameters and GFLOPs compared to the image-only baseline due to its dual-encoder design, a common trait in multimodal approaches. Compared to Attention Fusion, FreDFuse introduces only a small overhead: +2.56\% parameters, +0.46\% GFLOPs, but a higher latency (+24.26\%), mainly due to its pyramid-based decomposition, which adds memory and synchronization overhead despite low FLOP cost.



\section{Conclusion}

We present \textit{FUSE}, a framework for image-event joint depth estimation that tackles two challenges: cross-modal knowledge transfer and spectral mismatch during fusion. A two-stage adapter tuning transfers priors from image depth models to the joint domain without labeled image-event pairs. The proposed \texttt{FreDFuse} module leverages Gaussian-Laplacian pyramids to align frequency components from the two modalities. \textit{FUSE} achieves state-of-the-art performance on MVSEC and DENSE benchmarks.

However, \textit{FUSE} has several limitations. First, converting asynchronous events into fixed-interval frames reduces temporal fidelity, missing fine motion details. Second, it processes samples independently, ignoring inter-frame dynamics and temporal consistency. Finally, its relatively high latency—due to dual-stream processing and multiscale fusion—limits real-time deployment. Future work should focus on two main aspects. The first is to explore native asynchronous processing and recurrent architectures to better exploit temporal dynamics and improve depth consistency. The second is to optimize the fusion module and investigate model compression techniques, which are crucial for reducing latency and enabling real-time deployment.










\begin{table}[t]
\renewcommand\arraystretch{1.2}
\centering
\caption{Efficiency Comparisons Between \textit{FUSE} and Baselines.}
\label{tab:overhead}
\begin{adjustbox}{width=\columnwidth}
\begin{tabular}{c|ccc}
\Xhline{1.1pt}
Model & Paras (M) & GFLOPs & Latency (ms) \\ \hline
Image-Only Baseline & 335.32 & 309.61 & 29.20 \\
Attention Fusion & 660.05 & 549.21 & 57.90 \\
\textbf{\textit{FUSE}} & \textbf{676.92} & \textbf{551.74} & \textbf{71.95} \\
\Xhline{1.1pt}
\end{tabular}
\end{adjustbox}
\end{table}

\section*{ACKNOWLEDGMENT}

The research was supported by the National Natural Science Foundation of China (U23B2009, 62471158).

\printbibliography

@ARTICLE{chen2024frequency,
  title={Frequency-Aware Feature Fusion for Dense Image Prediction}, 
  author={Chen, Linwei and Fu, Ying and Gu, Lin and Yan, Chenggang and Harada, Tatsuya and Huang, Gao},
  journal={IEEE Transactions on Pattern Analysis and Machine Intelligence}, 
  volume={46},
  number={12},
  pages={10763-10780},
  year={2024},
  publisher={IEEE},
}

@article{eventsurvey,
  title={Event-based vision: A survey},
  author={Gallego, Guillermo and Delbr{\"u}ck, Tobi and Orchard, Garrick and Bartolozzi, Chiara and Taba, Brian and Censi, Andrea and Leutenegger, Stefan and Davison, Andrew J and Conradt, J{\"o}rg and Daniilidis, Kostas and others},
  journal={IEEE Transactions on Pattern Analysis and Machine Intelligence},
  volume={44},
  number={1},
  pages={154--180},
  year={2020},
  publisher={IEEE}
}

@article{midas,
  title={Towards robust monocular depth estimation: Mixing datasets for zero-shot cross-dataset transfer},
  author={Ranftl, Ren{\'e} and Lasinger, Katrin and Hafner, David and Schindler, Konrad and Koltun, Vladlen},
  journal={IEEE Transactions on Pattern Analysis and Machine Intelligence},
  volume={44},
  number={3},
  pages={1623--1637},
  year={2020},
  publisher={IEEE}
}

@ARTICLE{ereformer,
  author={Liu, Xu and Li, Jianing and Shi, Jinqiao and Fan, Xiaopeng and Tian, Yonghong and Zhao, Debin},
  journal={IEEE Transactions on Circuits and Systems for Video Technology}, 
  title={Event-Based Monocular Depth Estimation With Recurrent Transformers}, 
  year={2024},
  volume={34},
  number={8},
  pages={7417-7429},
}

@incollection{forgetcatastrophic,
title = {Catastrophic Interference in Connectionist Networks: The Sequential Learning Problem},
booktitle = {Psychology of Learning and Motivation},
publisher = {Academic Press},
volume = {24},
pages = {109-165},
year = {1989},
issn = {0079-7421},
author = {Michael McCloskey and Neal J. Cohen},
}

@INPROCEEDINGS{e2depth,
  author={Hidalgo-Carrió, Javier and Gehrig, Daniel and Scaramuzza, Davide},
  booktitle={2020 International Conference on 3D Vision (3DV)}, 
  title={Learning Monocular Dense Depth from Events}, 
  year={2020},
  volume={},
  number={},
  pages={534-542},
}

@INPROCEEDINGS{eventbad,
  author={Sabater, Alberto and Montesano, Luis and Murillo, Ana C.},
  booktitle={2022 IEEE/CVF Conference on Computer Vision and Pattern Recognition Workshops (CVPRW)}, 
  title={Event Transformer. A sparse-aware solution for efficient event data processing}, 
  year={2022},
  volume={},
  number={},
  pages={2676-2685},
}

@article{zhang2023cmx,
  title={CMX: Cross-Modal Fusion for RGB-X Semantic Segmentation With Transformers}, 
  author={Zhang, Jiaming and Liu, Huayao and Yang, Kailun and Hu, Xinxin and Liu, Ruiping and Stiefelhagen, Rainer},
  journal={IEEE Transactions on intelligent transportation systems},
  volume={24},
  number={12},
  pages={14679--14694},
  year={2023},
  publisher={IEEE}
}

@inproceedings{zhou2023rgb,
  title={RGB-Event Fusion for Moving Object Detection in Autonomous Driving}, 
  author={Zhou, Zhuyun and Wu, Zongwei and Boutteau, R{\'e}mi and Yang, Fan and Demonceaux, C{\'e}dric and Ginhac, Dominique},
  booktitle={2023 IEEE International Conference on Robotics and Automation (ICRA)},
  pages={7808--7815},
  year={2023},
  organization={IEEE}
}

@inproceedings{shi2023even,
  title={EVEN: An Event-Based Framework for Monocular Depth Estimation at Adverse Night Conditions}, 
  author={Shi, Peilun and Peng, Jiachuan and Qiu, Jianing and Ju, Xinwei and Lo, Frank Po Wen and Lo, Benny},
  booktitle={2023 IEEE International Conference on Robotics and Biomimetics (ROBIO)},
  pages={1--7},
  year={2023},
  organization={IEEE}
}

@inproceedings{hmnet,
  title={Hierarchical Neural Memory Network for Low Latency Event Processing}, 
  author={Hamaguchi, Ryuhei and Furukawa, Yasutaka and Onishi, Masaki and Sakurada, Ken},
  booktitle={2023 IEEE/CVF Conference on Computer Vision and Pattern Recognition (CVPR)}, 
  pages={22867--22876},
  year={2023}
}

@inproceedings{mixed,
  title={Improved Event-Based Dense Depth Estimation via Optical Flow Compensation}, 
  author={Shi, Dianxi and Jing, Luoxi and Li, Ruihao and Liu, Zhe and Wang, Lin and Xu, Huachi and Zhang, Yi},
  booktitle={2023 IEEE International Conference on Robotics and Automation (ICRA)},
  pages={4902--4908},
  year={2023},
  organization={IEEE}
}

@inproceedings{imgbad1,
  title={Structure-Centric Robust Monocular Depth Estimation via Knowledge Distillation},
  author={Chen, Runze and Luo, Haiyong and Zhao, Fang and Yu, Jingze and Jia, Yupeng and Wang, Juan and Ma, Xuepeng},
  booktitle={Proceedings of the Asian Conference on Computer Vision},
  pages={2970--2987},
  year={2024}
}

@inproceedings{srfnet,
  title={SRFNet: Monocular Depth Estimation with Fine-grained Structure via Spatial Reliability-oriented Fusion of Frames and Events}, 
  author={Pan, Tianbo and Cao, Zidong and Wang, Lin},
  booktitle={2024 IEEE International Conference on Robotics and Automation (ICRA)},
  pages={10695--10702},
  year={2024},
  organization={IEEE}
}

@article{ramnet,
  title={Combining Events and Frames Using Recurrent Asynchronous Multimodal Networks for Monocular Depth Prediction}, 
  author={Gehrig, Daniel and R{\"u}egg, Michelle and Gehrig, Mathias and Hidalgo-Carri{\'o}, Javier and Scaramuzza, Davide},
  journal={IEEE Robotics and Automation Letters},
  volume={6},
  number={2},
  pages={2822--2829},
  year={2021},
  publisher={IEEE}
}

@inproceedings{pcdepth,
  title={PCDepth: Pattern-based Complementary Learning for Monocular Depth Estimation by Best of Both Worlds},
  author={Liu, Haotian and Qu, Sanqing and Lu, Fan and Bu, Zongtao and Röhrbein, Florian and Knoll, Alois and Chen, Guang},
  booktitle={2024 IEEE/RSJ International Conference on Intelligent Robots and Systems (IROS)},
  pages={11187--11194},
  year={2024},
  organization={IEEE}
}

@inproceedings{depthanything1,
  title={Depth Anything: Unleashing the Power of Large-Scale Unlabeled Data}, 
  author={Yang, Lihe and Kang, Bingyi and Huang, Zilong and Xu, Xiaogang and Feng, Jiashi and Zhao, Hengshuang},
  booktitle={2024 IEEE/CVF Conference on Computer Vision and Pattern Recognition (CVPR)}, 
  pages={10371--10381},
  year={2024}
}

@inproceedings{depthanything2,
 author = {Yang, Lihe and Kang, Bingyi and Huang, Zilong and Zhao, Zhen and Xu, Xiaogang and Feng, Jiashi and Zhao, Hengshuang},
 booktitle = {Advances in Neural Information Processing Systems},
 pages = {21875--21911},
 publisher = {Curran Associates, Inc.},
 title = {Depth Anything V2},
 volume = {37},
 year = {2024}
}

@article{eigen,
  title={Depth Map Prediction from A Single Image Using A Multi-scale Deep Network},
  author={Eigen, David and Puhrsch, Christian and Fergus, Rob},
  journal={Advances in neural information processing systems},
  volume={27},
  year={2014}
}

@INPROCEEDINGS{adabins,
  author={Farooq Bhat, Shariq and Alhashim, Ibraheem and Wonka, Peter},
  booktitle={2021 IEEE/CVF Conference on Computer Vision and Pattern Recognition (CVPR)}, 
  title={AdaBins: Depth Estimation Using Adaptive Bins}, 
  year={2021},
  volume={},
  number={},
  pages={4008-4017},
}

@ARTICLE{binsformer,
  author={Li, Zhenyu and Wang, Xuyang and Liu, Xianming and Jiang, Junjun},
  journal={IEEE Transactions on Image Processing}, 
  title={BinsFormer: Revisiting Adaptive Bins for Monocular Depth Estimation}, 
  year={2024},
  volume={33},
  number={},
  pages={3964-3976},
}

@ARTICLE{nddepth,
  author={Shao, Shuwei and Pei, Zhongcai and Chen, Weihai and Chen, Peter C. Y. and Li, Zhengguo},
  journal={IEEE Transactions on Pattern Analysis and Machine Intelligence}, 
  title={NDDepth: Normal-Distance Assisted Monocular Depth Estimation and Completion}, 
  year={2024},
  volume={46},
  number={12},
  pages={8883-8899},
}

@INPROCEEDINGS{newcrf,
  author={Yuan, Weihao and Gu, Xiaodong and Dai, Zuozhuo and Zhu, Siyu and Tan, Ping},
  booktitle={2022 IEEE/CVF Conference on Computer Vision and Pattern Recognition (CVPR)}, 
  title={Neural Window Fully-connected CRFs for Monocular Depth Estimation}, 
  year={2022},
  volume={},
  number={},
  pages={3906-3915},
}

@INPROCEEDINGS{dpt,
  author={Ranftl, René and Bochkovskiy, Alexey and Koltun, Vladlen},
  booktitle={2021 IEEE/CVF International Conference on Computer Vision (ICCV)}, 
  title={Vision Transformers for Dense Prediction}, 
  year={2021},
  volume={},
  number={},
  pages={12159-12168},
}

@inproceedings{lora,
title={Lo{RA}: Low-Rank Adaptation of Large Language Models},
author={Edward J Hu and Yelong Shen and Phillip Wallis and Zeyuan Allen-Zhu and Yuanzhi Li and Shean Wang and Lu Wang and Weizhu Chen},
booktitle={International Conference on Learning Representations},
year={2022},
}

@INPROCEEDINGS{gehrig2019end,
  author={Gehrig, Daniel and Loquercio, Antonio and Derpanis, Konstantinos and Scaramuzza, Davide},
  booktitle={2019 IEEE/CVF International Conference on Computer Vision (ICCV)}, 
  title={End-to-End Learning of Representations for Asynchronous Event-Based Data}, 
  year={2019},
  volume={},
  number={},
  pages={5632-5642},
}

@ARTICLE{mvsec,
  author={Zhu, Alex Zihao and Thakur, Dinesh and Özaslan, Tolga and Pfrommer, Bernd and Kumar, Vijay and Daniilidis, Kostas},
  journal={IEEE Robotics and Automation Letters}, 
  title={The Multivehicle Stereo Event Camera Dataset: An Event Camera Dataset for 3D Perception}, 
  year={2018},
  volume={3},
  number={3},
  pages={2032-2039},
}

@inproceedings{nyuv2,
  title={Indoor Segmentation And Support Inference from RGBD Images},
  author={Silberman, Nathan and Hoiem, Derek and Kohli, Pushmeet and Fergus, Rob},
  booktitle={European Conference on Computer Vision},
  pages={746--760},
  year={2012},
  organization={Springer}
}

@article{kitti,
  author = {Andreas Geiger and Philip Lenz and Christoph Stiller and Raquel Urtasun},
  title = {Vision Meets Robotics: The KITTI Dataset},
  journal = {International Journal of Robotics Research},
  volume = {32},
  number = {11},
  pages = {1231-1237},
  year = {2013}
}

@inproceedings{yao2024improving,
  title={Improving Domain Generalization in Self-supervised Monocular Depth Estimation via Stabilized Adversarial Training},
  author={Yao, Yuanqi and Wu, Gang and Jiang, Kui and Liu, Siao and Kuai, Jian and Liu, Xianming and Jiang, Junjun},
  booktitle={European Conference on Computer Vision},
  pages={183--201},
  year={2024},
  organization={Springer}
}

@inproceedings{imgbag4,
  title={Think Small, Act Big: Primitive Prompt Learning for Lifelong Robot Manipulation},
  author={Yao, Yuanqi and Liu, Siao and Song, Haoming and Qu, Delin and Chen, Qizhi and Ding, Yan and Zhao, Bin and Wang, Zhigang and Li, Xuelong and Wang, Dong},
  booktitle={Proceedings of the Computer Vision and Pattern Recognition Conference},
  pages={22573--22583},
  year={2025}
}

@article{zhu2018ev,
  title={EV-FlowNet: Self-supervised optical flow estimation for event-based cameras},
  author={Zhu, Alex Zihao and Yuan, Liangzhe and Chaney, Kenneth and Daniilidis, Kostas},
  journal={arXiv preprint arXiv:1802.06898},
  year={2018}
}

@article{xu2024unveiling,
  title={Unveiling the depths: A multi-modal fusion framework for challenging scenarios},
  author={Xu, Jialei and Liu, Xianming and Jiang, Junjun and Jiang, Kui and Li, Rui and Cheng, Kai and Ji, Xiangyang},
  journal={arXiv preprint arXiv:2402.11826},
  year={2024}
}

\end{document}